\pdfoutput=1

\documentclass[11pt]{article}

\usepackage{nlpaics2026}

\usepackage{times}
\usepackage{amssymb} 
\usepackage{comment}
\usepackage{latexsym}
\usepackage[T1]{fontenc}
\usepackage[utf8]{inputenc}
\usepackage{microtype}
\usepackage{graphicx}
\usepackage{booktabs}
\usepackage{amsmath}
\usepackage{xcolor}
\title{Analysing Self-Harm Representations in Language Models: a Cross-Architecture Study}

\author{
  Luis Espinosa-Anke \quad Carla Perez-Almendros \\
  Cardiff University \\
  \texttt{\{espinosa-ankel, perezalmendrosc\}@cardiff.ac.uk}
}

\begin{document}
\maketitle

\begin{abstract}

Detecting self-harm is a challenging and high-stakes task requiring the highest accuracy to enable timely intervention and flagging at-risk users. In this paper, we present an analysis of how LLMs represent such self-harm posts.
We perform two experiments:
(1) We train and evaluate linear probes across all layers of each model on two self-harm datasets and find that self-harm information crystallizes in the final 3 - 7\% of network layers (93 to 97\% depth). (2) We extract contrastive self-harm directions and, after performing a normalization step, we find that the most accurate probes are not necessarily the most linearly separable. In particular, we find Gemma-3-4B to represent this \textit{contrastive self-harm direction} in a slightly different, more intricate way than the other LLMs \footnote{Code available at \url{https://github.com/luisespinosaanke/self-harm-analysis}.}. 
\end{abstract}

\section{Introduction}

Self-harm content on social media poses a significant public health concern. Researchers have focused on using AI for analyzing mental health disorders \cite{owen2024ai}, with many building classifiers for depression \cite{nadeem2016identifyingdepressiontwitter} and self-harm detection \citep{antypas2025sensitive, rozova2022detection, ayre2021developing}. Automated detection systems, thus, play a critical role in content moderation \citep{coppersmith2018natural, ji2022suicidal} and for flagging depressive disorders \cite{DBLP:conf/insci/LeivaF17}. Past approaches include using LSTMs \cite{yates-etal-2017-depression}, BERT-based transformers \cite{owen-etal-2020-towards} and LLMs \cite{ohse2024zero}. However, there is still a significant gap in understanding \emph{how} LLMs internally represent such content: where in the network self-harm concepts crystallize, how geometrically stable these representations are across layers, or whether the structure is consistent across architectures and scales. We argue that gaining deeper understanding on these questions can contribute dramatically to building safer systems that go beyond prompt engineering, and instead include explicit, testable components, which also enables more realistic intervention, governance and escalation \cite{reddy2025preventingtessamodularsafety}\footnote{When deployed without such features, AI-based assistants have proven brittle in real world scenarios, and caused genuine public health issues: \url{https://www.bbc.co.uk/news/world-us-canada-65771872}.}. In this context, ``representation engineering'' \citep{zou2023representation} offers a promising framework for understanding model internals through the lens of residual stream activations. In particular, the extraction of contrastive directions (typically via datasets of contrastive prompts that elicit opposed behaviours in LLMs) allows us to probe concept separability as well as to characterise the geometry of safety-relevant embeddings \citep{turner2023activation, li2024inference}, with direct applications in bias mitigation \citep{siddique2026shifting}. In this paper, thus, we ask ourselves the question whether self-harm \textit{content} is equally, more or less likely to be captured by a single direction, a proven feature in ``refusal'', for instance, a well-known proxy for embedded safety in AI-assistants \cite{arditi2024refusallanguagemodelsmediated}. We found consistent late-layer crystallisation of self-harm concepts, and a cross-layer block-diagonal embedding of contrastive directions, which we argue paves the way for future more effecrtive intervention.


\section{Related Work}

Automated detection of suicidal ideation and self-harm has long been studied \citep{DBLP:conf/insci/LeivaF17,coppersmith2018natural, ji2022suicidal, yang2023mentalllama,ghosh-etal-2025-just}. 
However, they are noteably unable to handle mental health-specific challenges like prioritising crisis intervention accuracy or preventing escalations \cite{zhang2025safetyrefusalreasoningenhancedfinetuning,lee2025programmingrefusalconditionalactivation,11269647,stamatis2026simulations20000realconversations,Lyu_2025}.
Seeking AI-safety, LLMs are reinforced with human feedback to replicate human preferences \cite{ziegler2019fine, bai2022training}. Initiatives like Constitutional AI \cite{bai2022constitutional} also allow models to learn from their own outputs to maximize safe and accurate responses. Another approach to safety involves the better understanding of LLMs behaviour through representation analysis, in particular leveraging linear probes 
\citep{alain2016understanding, belinkov2022probing, burns2023discovering}. Recent work has extended probing to latent knowledge discovery \citep{burns2023discovering} and safety-relevant concepts. 
 \citet{zou2023representation} 
 characterizes \emph{where} in the activation space the concept lives and how it evolves across layers, by extracting mean-difference contrastive directions. \citet{subramani2022extracting} and \citet{turner2023activation} showed that such directions can be used for inference-time control, while \citet{siddique2026shifting} and \citet{li2024inference} demonstrated downstream applications in bias mitigation and truthfulness.

\section{Data and Models}


We evaluate on two datasets. First, \textbf{X-Sensitive (X-S)} \citet{antypas2025sensitive}, a multi-label corpus of English-language social media posts annotated for six sensitive content categories: self-harm, conflictual, profanity, sex, drugs, and spam. We combine all available splits (train, validation, test) and extract 200 self-harm positive posts (\texttt{selfharm}$=1$) and the 200 control posts where all categories are 0. Second, \textbf{SH-Detection (SH-D).} \citet{tharsi2025selfharm}, where
self-harm posts have \texttt{class}$=$\texttt{``self-harm''}; all remaining posts serve as controls. Unlike X-S, controls may include posts from any non-self-harm category. We sample a balanced binary 600-post dataset.
For both datasets, posts shorter than 20 characters are discarded, URLs are removed, whitespace is collapsed, and @mentions are anonymised. Representative examples are shown in Table~\ref{tab:examples}. In terms of LLMs, we select four instruction-tuned transformer architectures: \textbf{Qwen3-0.6B} (0.6B parameters, 28 layers, $d_\text{model}=1024$; \citealp{qwen2025qwen3}), \textbf{Llama~3.2-1B-Instruct} (1B parameters, 16 layers, $d_\text{model}=2048$; \citealp{grattafiori2024llama}), \textbf{Llama~3.2-3B-Instruct} (3B parameters, 28 layers, $d_\text{model}=3072$; \citealp{grattafiori2024llama}), and \textbf{Gemma-3-4B-it} (4B parameters, 34 layers, $d_\text{model}=2560$; \citealp{team2025gemma}). 
We access internal representations via TransformerLens \citep{nanda2022transformerlens}, hooking into \texttt{hook\_resid\_post} at each transformer block to extract residual stream activations after the full sublayer stack (attention + MLP). We use the \emph{last token position} as the representation of each input, following standard practice for causal LMs where the final token aggregates full left context.

\begin{table}[t]
\centering
\small
\begin{tabular}{p{0.16\columnwidth}p{0.7\columnwidth}}
\toprule
\textbf{Source} & \textbf{Example post} \\
\midrule
X-S/S-H & \emph{``feeling suicidal retweet and like to stop me, comment to encourage me help''} \\
X-S/C   & \emph{``Just watched the game, incredible performance tonight''} \\ 
\hline
SH-D/S-H   & \emph{``i'm useless and a waste of space but at least i finally got the courage to off myself''} \\
SH-D/C   & \emph{``done with the stress''} \\

\bottomrule
\end{tabular}
\caption{Representative self-harm (S-H) and control (C) examples from X-Sensitive (X-S) and self-harm detection (SH-D). 
}
\label{tab:examples}
\end{table}


\section{Experiments}

\subsection{Layer-wise Probing}

\begin{figure}[]
\centering
\includegraphics[width=\columnwidth]{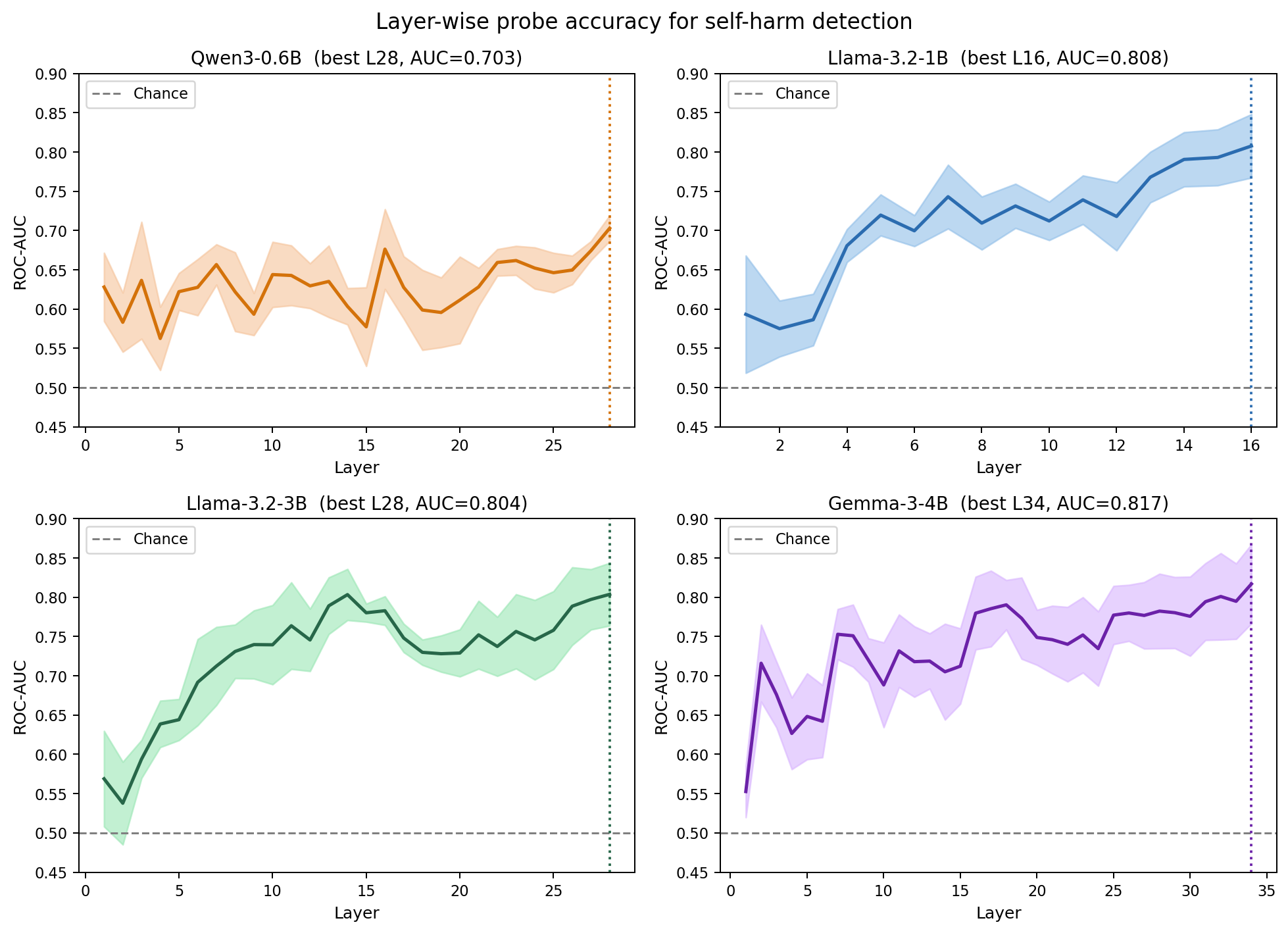}
\caption{Layer-wise probe ROC-AUC for all four models on X-Sensitive. 
Shaded regions denote $\pm$1 SD across 5-fold CV. 
}
\label{fig:probes}
\end{figure}

\begin{figure}[h!]
\centering
\includegraphics[width=\columnwidth]{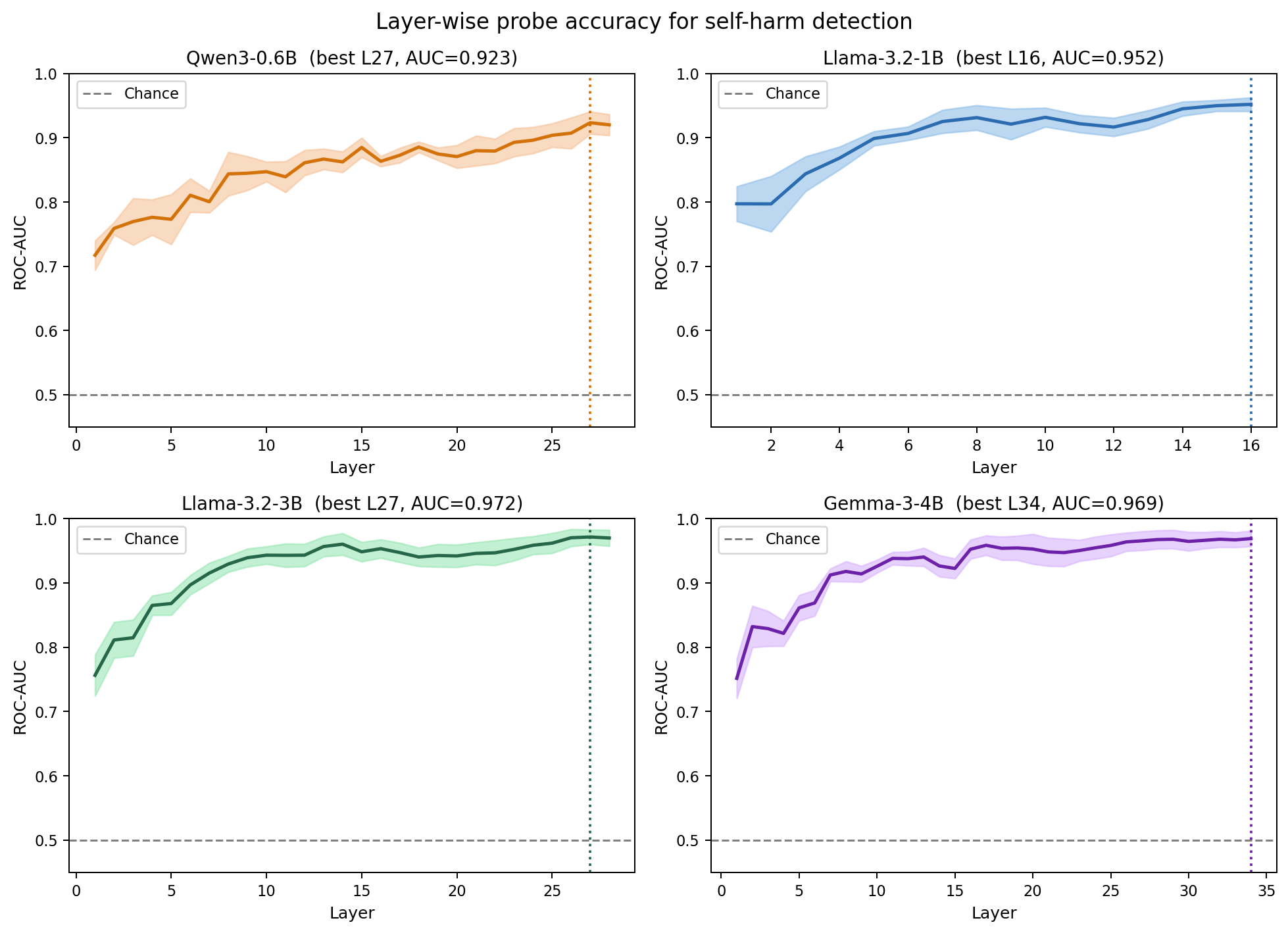}
\caption {Layer-wise probe ROC-AUC for all four models on SH-Detection. Shaded regions denote $\pm$1 SD across 5-fold CV.} 

\label{fig:probes_SH}
\end{figure}

\paragraph{Setup.} For each layer $\ell \in \{1, \ldots, L\}$, we extract last-token residual-stream activations $\mathbf{a}_\ell \in \mathbb{R}^{d_\text{model}}$ for all posts in each dataset (400 for X-S, 600 for SH-D) and train an $\ell_2$-regularised logistic regression probe ($C=1.0$, \texttt{lbfgs} solver, max 1000 iterations) using 5-fold stratified cross-validation. We report mean ROC-AUC as our primary metric, as it measures discriminability independently of classification threshold and is robust to class balance effects. Activations are not normalised prior to probing.

\paragraph{Results.} Table~\ref{tab:probe} summarises the probing results. Across all four models and both datasets, self-harm information crystallizes at 93 - 97\% of network depth. 
In terms of classification, AUC ranges from 0.703 (Qwen3-0.6B) to 0.817 (Gemma-3-4B) in X-S. On SH-D, all models improve substantially: AUC ranges from 0.923 (Qwen3-0.6B) to 0.972 (Llama~3.2-3B), with significantly lower variance (standard deviation halves). 
Figures ~\ref{fig:probes} and \ref{fig:probes_SH} show the full layer-wise increasing AUC curves for X-S and SH-Detection, respectively. As for a dataset comparison, SH-D achieves uniformly higher AUC with lower variance. We attribute this to corpus differences, e.g., presence of figurative or metaphorical content in X-S's vs. SH-D's cleaner binary labelling. Such figurative language also makes X-S's negative examples subtler, which might reduce probe ceiling performance.
\begin{table}[t]
\centering
\small
\resizebox{\columnwidth}{!}{%
\begin{tabular}{lccccc}
\toprule
 & & \multicolumn{2}{c}{\textbf{X-Sensitive}} & \multicolumn{2}{c}{\textbf{SH-Detection}} \\
\cmidrule(lr){3-4}\cmidrule(lr){5-6}
\textbf{Model} & \textbf{$L$} & \textbf{Best $\ell$} & \textbf{AUC} & \textbf{Best $\ell$} & \textbf{AUC} \\
\midrule
Qwen3-0.6B   & 28 & 28 & .703 $\pm$ .016 & 27 & .923 $\pm$ .017 \\
Llama~3.2-1B & 16 & 16 & .808 $\pm$ .040 & 16 & .952 $\pm$ .011 \\
Llama~3.2-3B & 28 & 28 & .804 $\pm$ .040 & 27 & .972 $\pm$ .011 \\
Gemma-3-4B   & 34 & 34 & .817 $\pm$ .050 & 34 & .969 $\pm$ .012 \\
\bottomrule
\end{tabular}
}
\caption{Best linear probe ROC-AUC per dataset. $L$ = total layers; Best $\ell$ is the layer maximising mean ROC-AUC across 5-fold CV.}
\label{tab:probe}
\end{table}

\paragraph{Error analysis.} At the best probe layer on X-S, false negatives mostly consist of self-harm posts using figurative or metaphorical language, legal or news contexts, or pop-culture references. False positives, on the other hand, are predominantly mental health advocacy posts, content-warning-prefixed posts, and third-person crisis descriptions, which use vocabulary related to critical situations, but do not convey first-person distress signals, which makes them legitimate confounders.

\subsection{Contrastive Direction Analysis}
\label{sec:geometry}

\paragraph{Setup.} Following \citet{zou2023representation}, we compute the contrastive self-harm direction at each layer:
\begin{equation}
\mathbf{d}_\ell = \frac{\bar{\mathbf{a}}_\ell^{+} - \bar{\mathbf{a}}_\ell^{-}}{\|\bar{\mathbf{a}}_\ell^{+} - \bar{\mathbf{a}}_\ell^{-}\|}
\end{equation}
where $\bar{\mathbf{a}}_\ell^{+}$ and $\bar{\mathbf{a}}_\ell^{-}$ are mean last-token activations of self-harm and control posts. Each example's scalar projection $s_i = \mathbf{a}_{\ell,i} \cdot \mathbf{d}_\ell$ provides a single-dimensional self-harm score. To enable cross-model comparison despite differences in activation magnitudes, we compute Cohen's~$d$ at the best-separation layer:
\begin{equation}
d = \frac{\bar{s}^{+} - \bar{s}^{-}}{\sqrt{(\sigma_{+}^2 + \sigma_{-}^2)/2}}
\end{equation}
We assess directional stability via the $L \times L$ pairwise cosine similarity matrix $\text{cos}(\mathbf{d}_i, \mathbf{d}_j)$.

\paragraph{Separability.} Table~\ref{tab:geometry} shows contrastive geometry results. The reason for normalizing with Cohen's~$d$ is that raw separation scores vary by orders of magnitude across models and datasets (e.g.\ 3.1 for Llama-1B on X-S vs.\ 6,767 for Gemma-3-4B on SH-D) and so comparison between models would be challenging.


\begin{table}[!t]
\centering
\small
\resizebox{0.95\columnwidth}{!}{
\begin{tabular}{lcccc}
\toprule
 & \multicolumn{2}{c}{\textbf{X-Sensitive}} & \multicolumn{2}{c}{\textbf{SH-Detection}} \\
\cmidrule(lr){2-3}\cmidrule(lr){4-5}
\textbf{Model} & \textbf{Best $\ell$} & \textbf{Cohen's $d$} & \textbf{Best $\ell$} & \textbf{Cohen's $d$} \\
\midrule
Qwen3-0.6B   & 28 & 0.95 & 27 & 2.03 \\
Llama~3.2-1B & 16 & 1.26 & 16 & 2.06 \\
Llama~3.2-3B & 28 & 1.14 & 27 & 2.08 \\
Gemma-3-4B   & 34 & 0.71 & 34 & 1.18 \\
\bottomrule
\end{tabular}
}
\caption{Contrastive direction Cohen's~$d$ at each model's best probe layer (from Table~\ref{tab:probe}). 
}
\label{tab:geometry}
\end{table}

\begin{figure}[!h]
\centering
\includegraphics[width=\columnwidth]{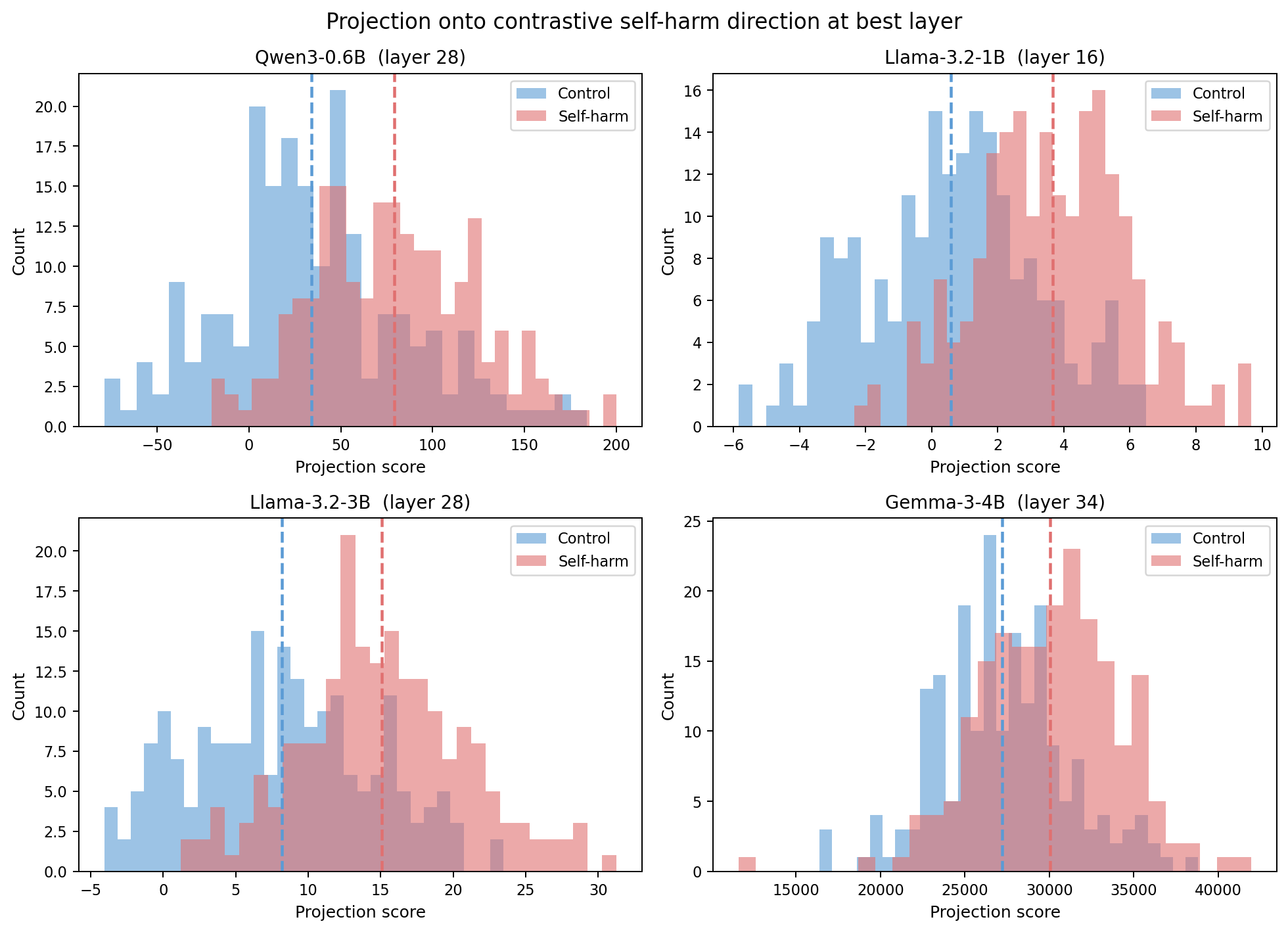}
\caption{Scalar projections onto the contrastive self-harm direction at each model's best layer for X-S. 
}
\label{fig:projection}
\end{figure}

\begin{figure}[t]
\centering
\includegraphics[width=\columnwidth]{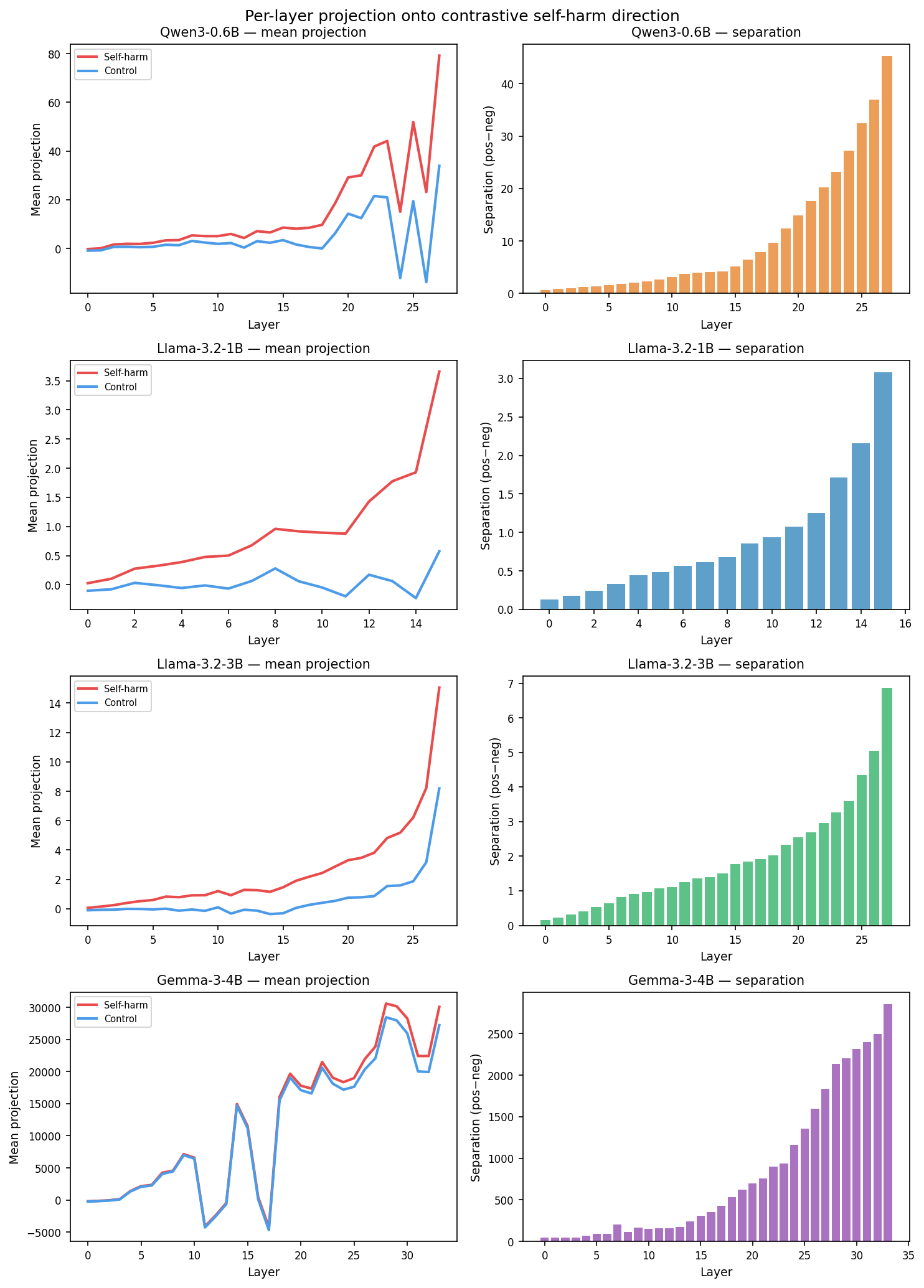}
\caption{Per-layer mean projection onto the contrastive self-harm direction for all four models (X-Sensitive). 
}
\label{fig:projection_layers}
\end{figure}

\begin{table}[t]
\centering
\small
\begin{tabular}{lcccc}
\toprule
 & \multicolumn{2}{c}{\textbf{X-Sensitive}} & \multicolumn{2}{c}{\textbf{SH-Detection}} \\
\cmidrule(lr){2-3}\cmidrule(lr){4-5}
\textbf{Model} & \textbf{Adj.} & \textbf{Cross} & \textbf{Adj.} & \textbf{Cross} \\
\midrule
Qwen3-0.6B   & 0.894 & 0.289 & 0.908 & 0.376 \\
Llama~3.2-1B & 0.758 & 0.222 & 0.801 & 0.273 \\
Llama~3.2-3B & 0.858 & 0.218 & 0.879 & 0.267 \\
Gemma-3-4B   & 0.897 & 0.239 & 0.918 & 0.135 \\
\bottomrule
\end{tabular}
\caption{Directional stability summary. \textbf{Adj.}: mean cosine similarity between adjacent-layer direction pairs. \textbf{Cross}: mean cosine between directions in the first and last thirds of the network. Full $L{\times}L$ matrices are in Figure~\ref{fig:cosines} (Appendix~\ref{app:cosines}).}
\label{tab:cosines}
\end{table}

On X-S, effect sizes range from 0.71 (Gemma-3-4B) to 1.26 (Llama~3.2-1B), all indicating large-effect separability. On SH-D, effect sizes roughly double: 1.18 to 2.08, with Gemma-3-4B again the lowest despite its superior probe AUC. This discrepancy, namely the high classification accuracy vs. a lower geometric separation for Gemma is consistent for both datasets. We argue that this points to a model-specific representational property rather than an artifact in the dataset. To give a concrete example, on SH-D the contrastive direction for Llama~3.2-3B peaks in Cohen's~$d$ at layer~14 ($d=2.20$), well before the final layers, while probe AUC continues rising to layer~27. This divergence sharpens the interpretation that probing and directional analysis capture complementary aspects: mid-network directions may be potent but not yet sufficient for accurate linear classification.

One geometric interpretation of Gemma's dissociation is as follows. A linear probe has access to the full $d_\text{model}$-dimensional activation space and can find a separating hyperplane exploiting any combination of features, including signal distributed, even if weakly, across many dimensions. Cohen's~$d$ on the mean-difference direction reveals whether the \emph{mean class centroids} are far apart relative to within-class spread along that single axis. High AUC with low $d$ therefore implies that Gemma's self-harm signal is not concentrated in one direction, but spread across the activation space. 
Next, we generate projection histograms at each model's best layer (Figure~\ref{fig:projection}), as well as per-layer mean projection curves (Figure~\ref{fig:projection_layers}) for X-S\footnote{Similar pattern for SH-D.}. These visuals show, first, a clear bimodal distribution with overlapping tails, which illustrates the geometric difference between the encodings of this dataset across models (although smaller for Gemma-3-4B). Also, that control means diverge mostly in the final two to four layers.

\paragraph{Directional stability.} Table~\ref{tab:cosines} summarises the block-diagonal structure visible in the full cosine matrices (Figure~\ref{fig:cosines}, Appendix~\ref{app:cosines}). On both corpora, adjacent-layer directions are highly aligned (0.76 to 0.92), while directions from the first and last thirds of each network are nearly orthogonal (between 0.14 and 0.38). This means that the self-harm concept is dynamically \emph{ re-represented} across depth, i.e., the direction \emph{rotates} substantially between early and late layers, and this structural property holds on the two datasets. Gemma-3-4B shows the lowest cross-comparison cosine on SH-D (0.135). 

\section{Conclusion}

We have presented a cross-architecture analysis of self-harm representations. In our experiments, self-harm concepts crystallize consistently in the 3-7\% ``latest'' layer block. Moreover, the self-harm contrastive direction is not stable across network depth, a direction extracted at an early layer does not transfer to late layers. In addition, the most accurate linear probes are not the most geometrically separable. Finally, the uniform performance gap between X-Sensitive and SH-Detection (0.12 to 0.17 AUC, $\sim$1 Cohen's~$d$ unit) also suggests that labelling quality and corpus composition have a large, systematic effect on representational quality.
For future work, we could extend these experiments to larger (7B up to or beyond 70B) models and to other sensitive content categories.

\bibliography{custom}
\bibliographystyle{acl_natbib}

\appendix

\section{Cross-layer Directional Stability}
\label{app:cosines}

\begin{figure*}[h]
\centering
\includegraphics[width=\textwidth]{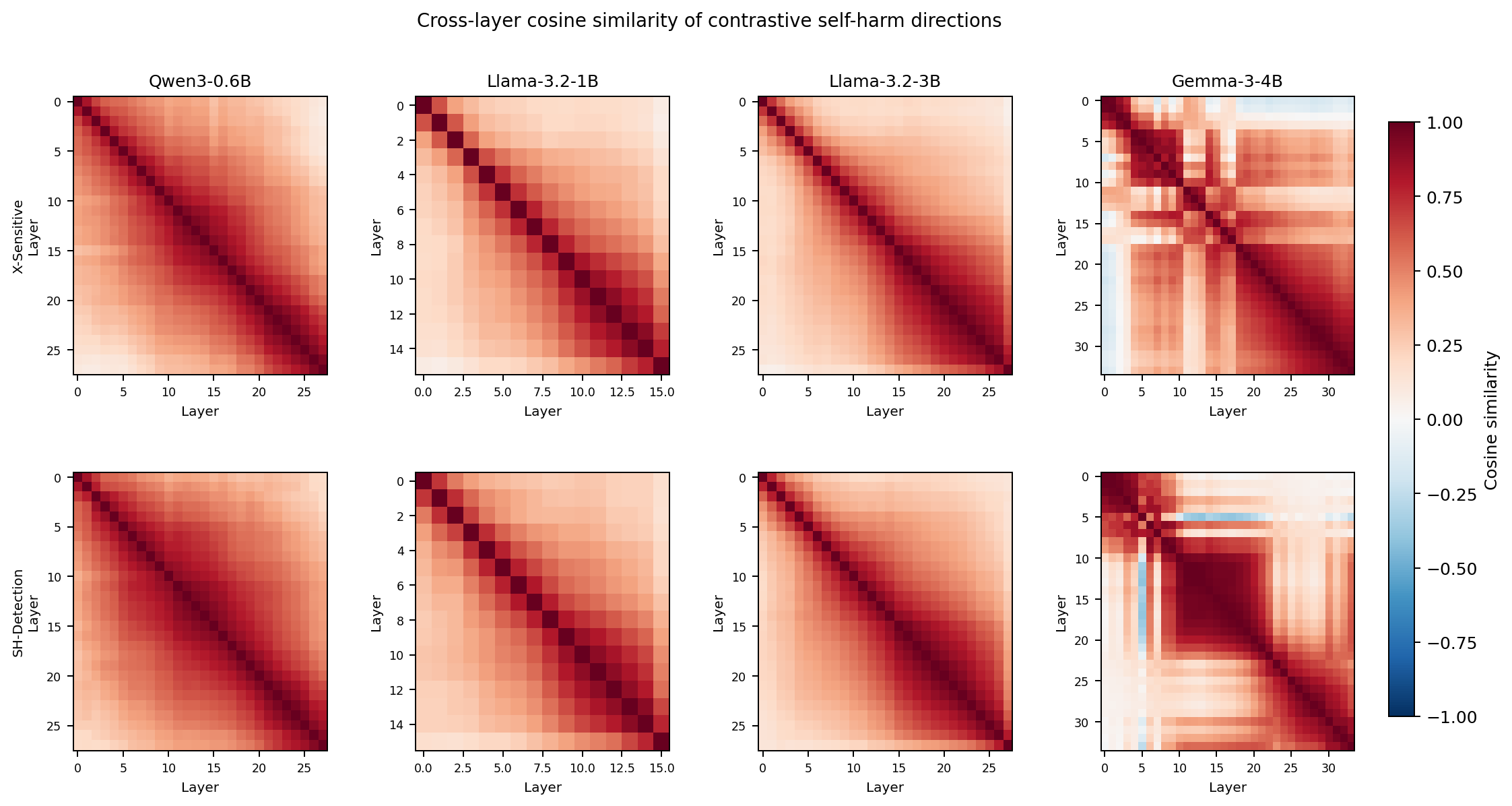}
\caption{Cross-layer cosine similarity of contrastive self-harm directions ($L \times L$ matrices) for X-Sensitive (top row) and SH-Detection (bottom row). High within-comparison and near-orthogonal cross-comparison cosines confirm the block-diagonal structure across all four models and both corpora.}
\label{fig:cosines}
\end{figure*}

Figure~\ref{fig:cosines} shows the full $L \times L$ pairwise cosine similarity matrices between contrastive self-harm directions across all layers, for all four models on both datasets. The block-diagonal structure is remarkably consistent across all plots: directions within the same network phase (early or late) are highly aligned, while directions from opposite phases are nearly orthogonal. Gemma-3-4B (rightmost column) stands out with an abrupt near-zero or negative cosine stripe at layers 5 through 8, particularly on SH-Detection, indicating a sharper directional transition than the other models. Summary statistics are reported in Table~\ref{tab:cosines}.

\end{document}